\documentclass[acmtog, nonacm]{acmart}

\setcopyright{none}
\renewcommand\footnotetextcopyrightpermission[1]{}

\AtBeginDocument{%
  \providecommand\BibTeX{{%
    \normalfont B\kern-0.5em{\scshape i\kern-0.25em b}\kern-0.8em\TeX}}}

\definecolor{amber}{rgb}{1.0, 0.3, 0}
\definecolor{green}{rgb}{0, 0.7, 0}

\newcommand{\zt}[1]{{\color{black}{#1}}}

\newcommand{\zy}[1]{{\color{black}{#1}}}

\newcommand{\reffig}[1]{{Fig.~\ref{fig:#1}}}

\newcommand{\refsec}[1]{{Sec.~\ref{sec:#1}}}
\newcommand{\refeqn}[1]{{Eqn.~(\ref{eqn:#1})}}

\newcommand{\eg}[1]{{\textit{e.g.,~}}}
\newcommand{\ie}[1]{{\textit{i.e.,~}}}

\usepackage{booktabs}
\usepackage[linesnumbered, ruled]{algorithm2e} 
\usepackage{graphicx}
\usepackage{float}
\usepackage{stfloats}
\usepackage{amsmath}
\usepackage{makecell}
\usepackage{comment}
\usepackage{subcaption}
\usepackage{hyperref}

\SetAlFnt{\small}
\SetAlCapFnt{\small}
\SetAlCapNameFnt{\small}
\SetAlCapHSkip{0pt}

\begin{document}
\title{Continuous Layout Editing of Single Images with Diffusion Models}

\author{Zhiyuan Zhang}
\email{zzhang452-c@my.cityu.edu.hk}
\affiliation{
    \institution{City University of Hong Kong}
    \city{Hong Kong SAR}
    \country{China}
}
\authornote{\label{equal}Both authors contributed equally to this research.}

\author{Zhitong Huang}
\email{luckyhzt@gmail.com}
\affiliation{
    \institution{City University of Hong Kong}
    \city{Hong Kong SAR}
    \country{China}
}
\authornotemark[1]

\author{Jing Liao}
\email{jingliao@cityu.edu.hk}
\affiliation{
    \institution{City University of Hong Kong}
    \city{Hong Kong SAR}
    \country{China}
}
\authornote{Corresponding author.}





\begin{abstract}
Recent advancements in large-scale text-to-image diffusion models have enabled many applications in image editing. However, none of these methods have been able to edit the layout of single existing images. To address this gap, we propose the first framework for layout editing of a single image while preserving its visual properties, thus allowing for continuous editing on a single image. Our approach is achieved through two key modules. First, to preserve the characteristics of multiple objects within an image, we disentangle the concepts of different objects and embed them into separate textual tokens using a novel method called masked textual inversion. Next, we propose a training-free optimization method to perform layout control for a pre-trained diffusion model, which allows us to regenerate images with learned concepts and align them with user-specified layouts. As the first framework to edit the layout of existing images, we demonstrate that our method is effective and outperforms other baselines that were modified to support this task. Our code will be freely available for public use upon acceptance at \textit{\textcolor{blue}{ \url{https://bestzzhang.github.io/continuous-layout-editing}}}.
\end{abstract}

\begin{teaserfigure}
    \centering
    \includegraphics[width=1.0\textwidth]{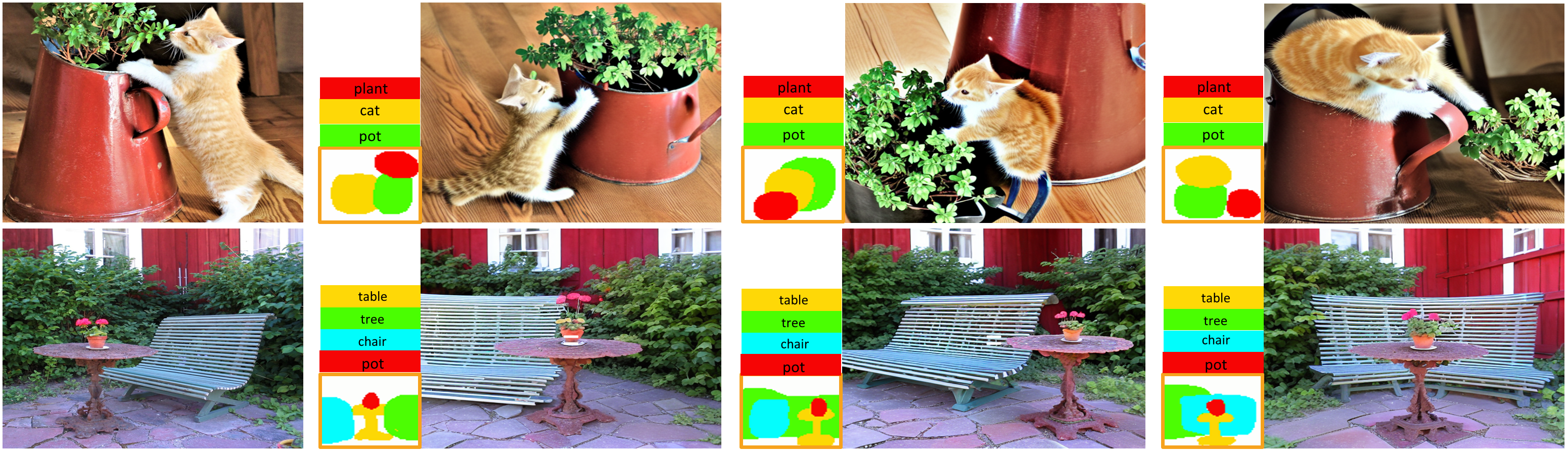}
    \caption{Our method can continuously edit the layout of a single image with multiple objects. The first column shows the input image, followed by three columns of edited results. The edited results preserve the visual properties of the input image while remaining faithful to the layout map provided by the user.}
    \label{fig:teaser}
\end{teaserfigure}

\maketitle

\section{Introduction}


Recent advances in text-to-image generation have made significant progress through the use of diffusion models trained on large-scale datasets~\cite{nichol_glide_2022,rombach_high-resolution_2022, saharia_photorealistic_nodate}. However, due to the inherent ambiguity of text and its limitations in expressing precise spatial relationships in the image space, controlling the layout of generated images is still a challenge for these large-scale text-to-image models. To address this issue, some latest methods have been proposed to enable layout control in image generation. These methods are typically based on pre-trained diffusion models, which either incorporate layout guidance as a new condition through fine-tuning \cite{avrahami_spatext_2023, zhang_adding_2023, li_gligen_2023} or optimize the noise diffusion process on-the-fly to achieve layout control \cite{bar-tal_multidiffusion_2023}.


Despite the success of existing layout control methods \cite{avrahami_spatext_2023, zhang_adding_2023, li_gligen_2023,bar-tal_multidiffusion_2023} in generating new images with controlled layouts, they are unable to rearrange and edit the layout of existing images. In practice, users may want to continuously edit the positions of objects in an existing image without altering its visual properties. For example, as illustrated in the first example of Figure \ref{fig:teaser}, a user may want to experiment with different layout options to find the best arrangement of a cat and a pot in an image. However, previous methods do not support this functionality since their layout control does not take into account the input image, and a new image with different cat and pot will be generated for each specified layout. To fill this gap, we propose the first framework for continuous layout editing of single input images.


One of the key challenges in continuous layout editing is preserving the visual properties of the input image, which requires  learning concepts for multiple objects within a single image and using the learned concepts to regenerate new images under different layouts. While some pioneer textual inversion methods~\cite{gal_image_2022, ruiz_dreambooth_2023} have proposed fine-tuning a text token embedding of pre-trained text-to-image diffusion models to learn the concept of an object from multiple images containing the same object, they are limited in their ability to learn multiple objects within a single image. To overcome this limitation, we propose a novel approach, called \textbf{masked textual inversion}, that disentangles the concepts of different objects within a single image and embeds them into separate tokens. By adding masks to the regions of each object, our method ensures that the visual characteristics of each object are effectively learned by the corresponding token embedding.


After learning the concepts of multiple objects within a single image, the next challenge is to control the positions of these objects to align with the desired layout. P2P~\cite{hertz_prompt--prompt_2022} suggests that the cross-attention of a pretrained text-to-image diffusion model can represent the position of the generated object associated with the corresponding text token and Attend-and-Excite~\cite{chefer2023attend} further utilizes the cross-attention to ensure the generation of objects. Inspired by these papers, we propose a novel, \textbf{training-free layout editing} method that iteratively optimizes the cross-attention during the diffusion process. This optimization is guided by a region loss that prioritizes the alignment of the specified object with its designated region in the layout by encouraging higher cross-attention between the object's text embedding and its corresponding region than with any other region in the image. Our approach enables precise and flexible control over the positions of objects in the image, without requiring additional training or fine-tuning of the pre-trained diffusion model.

Extensive experiments and perceptual studies have demonstrated that our proposed method is effective in editing the layout of single images and outperforms other baseline methods (modified to perform this task). We also provide a user interface for interactive layout editing to assist in the design process. In summary, our contributions to the field are as follows:
\begin{itemize}
    \item We propose the first framework which supports continuous layout editing of single images.
    \item We present a masked textual inversion method to learn disentangled concepts of multiple objects within single images.
    \item We propose a training-free optimization method to perform layout control with diffusion models.
\end{itemize}

\begin{figure*}[tbp]
    \centering
    \includegraphics[width=1.0\textwidth]{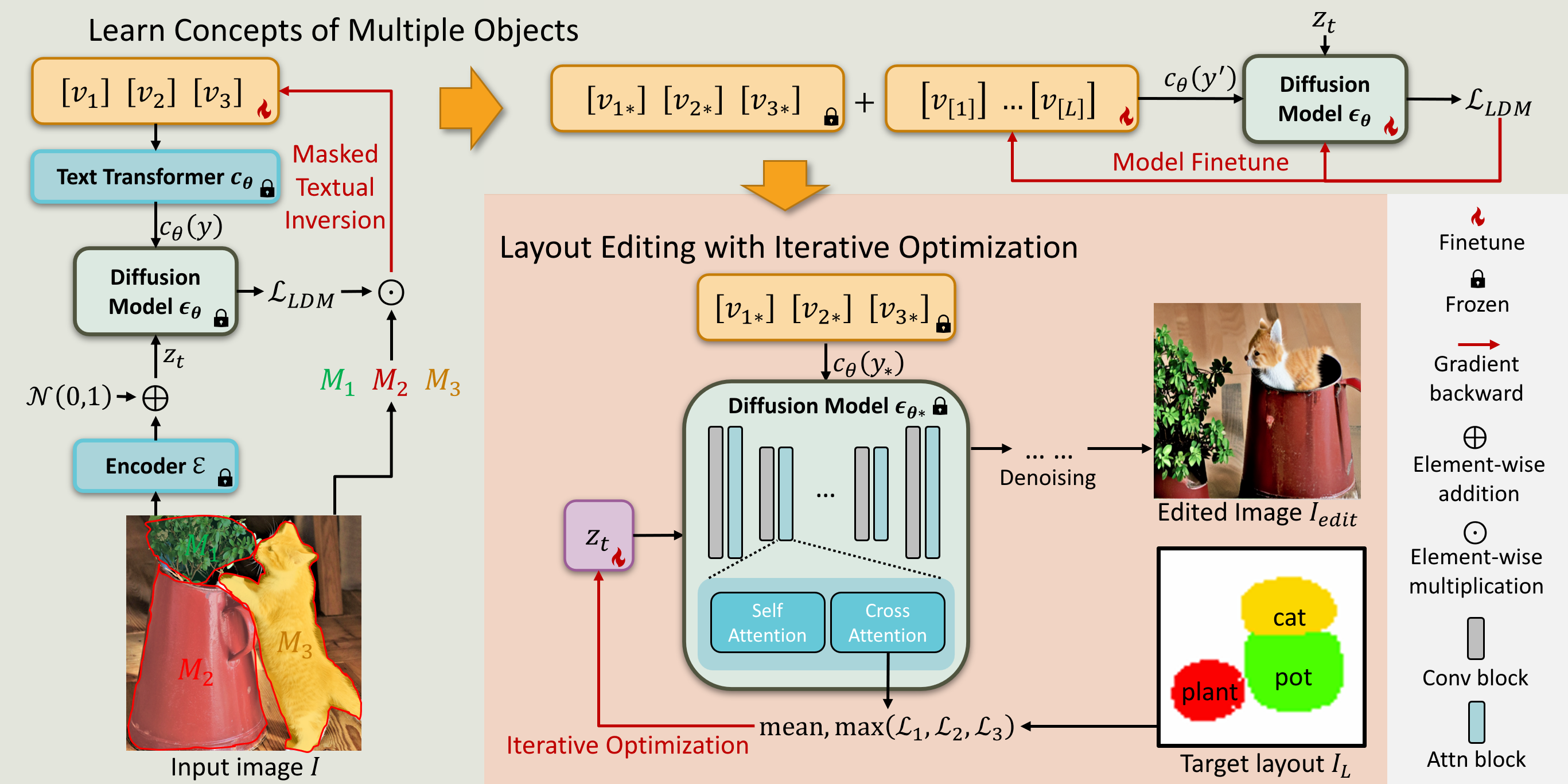}
    \caption{Overall framework of our method.}
    \label{fig:overall}
\end{figure*}

\section{Related Works}
In this section, we will review works related to our method, which includes diffusion models, image editing with diffusion models through textual inversion, and layout control with diffusion models.

\subsection{Diffusion Models}
Diffusion models have become one of the most popular generative models due to their impressive quality in image generation. The original DDPM \cite{ho_denoising_2020} simulates a Markovian process where Gaussian noise is added to clean images $x_0$ to create the noisy image $x_t$ in the forward process. Then, a model is trained to predict and remove the noise in $x_t$ to generate images. To accelerate the denoising process, DDIM \cite{song_denoising_2022} converts DDPM into a non-Markovian process, which requires no additional training.

Recently, text-to-image diffusion models \cite{balaji_ediff-i_2023, saharia_photorealistic_nodate, ramesh2022hierarchical} trained on large-scale datasets have gained significant attention due to their ability to generate diverse high-quality images with text prompts. Among them, Stable Diffusion \cite{rombach_high-resolution_2022} operates the diffusion process in latent space instead of pixel space, allowing it to generate high-resolution images.


\subsection{Image Editing with Textual Inversion}
By leveraging the power of pretrained text-to-image diffusion models, many image editing methods have been derived. Among them, a major category is to learn the concepts of objects or styles into text tokens and then generate new images with the extracted concepts. Pioneer works in this category include Textual inversion\cite{gal_image_2022} and DreamBooth\cite{ruiz_dreambooth_2023}. Textual inversion~\cite{gal_image_2022} embeds concepts of objects into pseudo-words, while DreamBooth~\cite{ruiz_dreambooth_2023} further finetunes the UNet to learn more details. However, these two methods can only extract a single common concept from multiple images. Multi-Concept~\cite{bar-tal_multidiffusion_2023} extends to learn multiple objects and explores the most effective layers in UNet to be finetuned; however, each concept still needs to be learned from multiple images. Overall, a method that can learn multiple concepts from a single image is still under exploration, and our masked textual inversion fills this gap.

\subsection{Layout Control with Diffusion Models}

Due to the sparsity and ambiguity of text descriptions, it is difficult to precisely control the layout of generated images by pretrained text-to-image diffusion models. To address this limitation, some layout control methods \cite{nichol_glide_2022,rombach_high-resolution_2022, saharia_photorealistic_nodate} based on diffusion models have been proposed, which can be divided into two categories.

The first category requires finetuning the pretrained text-to-image model to incorporate layout guidance as an extra condition besides text. Spatext~\cite{avrahami_spatext_2023} proposes to convert the concept of each object into CLIP image features with unCLIP~\cite{ramesh2022hierarchical}, which are then stacked at the target positions of that object to form a spatio-textual representation. This layout condition is concatenated with the noisy latent to control the layout during the denoising process. ControlNet~\cite{zhang_adding_2023} inserts additional conditions, such as the semantic maps for layout control, by utilizing a trainable copy of the original UNet model. The conditional copy and the original model are fused in intermediate layers to generate a conditioned output. GLIGEN~\cite{li_gligen_2023} adds additional trainable gated self-attention layers that take the information of layout conditions (i.e., bounding boxes) to control the layout of the generated image. A common limitation of these methods is that additional modules are added to the original UNet, and datasets of paired data (e.g., images and corresponding semantic maps) are required to finetune the diffusion model and added modules. Also, their capabilities are restricted by the training data.

The second category explores training-free layout control with on-the-flight optimization. The representative work is MultiDiffusion~\cite{bar-tal_multidiffusion_2023}, which denoises different crops of each object locally and then fuses the results globally for each denoising step. \zt{Compared with computing multiple denoising directions for each object, which may cause artifacts and discontinuities at the boundary of objects, our training-free layout editing method directly denoises the whole image and optimizes the image latent for layout control, to avoid the gaps and discontinuities among multiple denoising directions.}

Both categories of methods still focus on the layout control of generated images, which cannot be used to edit the layout of existing images. Our proposed framework targets this gap.

\section{Methodology}


\textbf{Preliminaries.} Our method is implemented with Stable Diffusion~\cite{rombach_high-resolution_2022}, a large-scale text-to-image model. Therefore, before discussing our method, we first introduce Latent Diffusion Models (LDMs)~\cite{rombach_high-resolution_2022}, which is the theory of Stable Diffusion. LDMs consist of two key stages. In the first stage, the encoder of an autoencoder maps the image to latent space $z_0 = \mathcal{E}(I)$, and a decoder maps it back to image $\mathcal{D}(\mathcal{E}(I)) \approx I$. In the second stage, a diffusion model $\epsilon_\theta$ is trained to denoise the noised latent $z_t = \sqrt{\alpha_t} z_0 + \sqrt{1-\alpha_t} \epsilon$, where $\alpha_t$ is a factor to determine noise level for each timestep $t$, and $\epsilon \sim \mathcal{N}(0, 1)$ is Gaussian noise. Then the diffusion model is trained to predict the added Gaussian noise with the LDM loss~\cite{ho_denoising_2020}:
\begin{equation}
    \mathcal{L}_{LDM} := \mathbb{E}_{z_0 \in \mathcal{E}(I), y, \epsilon \sim \mathcal{N}(0, 1), t} [ \| \epsilon - \epsilon_\theta(z_t, t, c_\theta(y)) \|^2_2 ]    \label{eqn:ldm}
\end{equation}
where $y$ is the input text condition, and $c_\theta$ is the text encoder.

\textbf{Overview.} As shown in \reffig{overall}, our methods can be divided into two stages. In the first stage, we learn the concepts of multiple objects from a single input image $I$ into text tokens $v_1, v_2, ..., v_N$ with masked textual inversion, where the regions of each object are specified by masks $M_1, M_2, ..., M_N$. We further learn the details of the objects by finetuning the diffusion model $\epsilon_\theta$ and optimizing the appended text tokens $v_{[1]},...,v_{[L]}$. 
After the first stage, we get the optimized text tokens for objects $y_* = [v_{1*},...,v_{N*}]$ and the finetuned model $\epsilon_{\theta*}$. 
Then, in the second stage, we rearrange the positions of the objects according to the user-specified layout map $I_L$ through a training-free layout editing method with optimization:
\begin{equation}
    I_{edit} = \texttt{Layout-control} (I, c_\theta(y_*), \epsilon_{\theta*}, I_L)
\end{equation}

\subsection{Learn Concepts of Multiple Objects within Single Image}
\label{sec:mask_text_inv}
To rearrange the layout of an input image, we first need to extract the concepts of multiple objects within the single input image to best preserve their visual characteristics, such as shape, color, and texture. We propose using masked textual inversion to learn and embed the concept of each individual object into a unique text token. We then fine-tune the diffusion model to better grasp the detailed texture of the learned objects.

\subsubsection{Masked textual inversion}


\zt{Original technique of text inversion only supports learning the concept of a single object from a set of images (typically 3-5). However, in our applications, we need to learn multiple concepts from a single image. As observed in \cite{avrahami_spatext_2023}, the latent vector $z_0=\mathcal{E}(I)$ encoded from the input image with the autoencoder \cite{rombach_high-resolution_2022} has local property in the spatial dimension and the encoder performs like a down-sampler. Therefore, we can disentangle the concepts of different objects by simply applying a spatial mask. Instead of calculating the loss of the whole latent, we only propagate the loss within the region of the object to update the corresponding text token:
\begin{align}
    v_{k*} = \mathop{\arg \min}\limits_{v_k} \: \mathbb{E}_{z_0 \in \mathcal{E}(I), y, \epsilon \sim \mathcal{N}(0, 1), t} &     \nonumber
    \\
    [ M_k \odot \| \epsilon - & \epsilon_\theta(z_t, t, c_\theta(y)) \|^2_2 ]    \label{eqn:opt_text}
\end{align}
where $M_k$ is the mask of the $k$-th object ($k = 1, ..., N$ is the index of the $N$ objects), $v_k$ is the corresponding text token of the $k$-th object, and the input text condition $y$ consists of text tokens $[v_1, ..., v_N]$. The mask can either be generated coarsely by hand or automatically using CLIP Segmentation~\cite{luddecke2022image}. We repeat this process independently for each of the $N$ objects to optimize the text tokens for each object. To avoid overfitting, we only run each optimization for 200 steps, which is much less than the original textual inversion 3000-5000 steps).
}

\subsubsection{Model fintuning}
\zt{
A single text token can only store limited information of an object, which may cause obvious distortion or artifacts during sampling. Therefore, we propose further fine-tuning the denoising network $\epsilon_\theta$ to better grasp the detailed texture of the objects. In multi-concept customization \cite{kumari_multi-concept_2022}, it was discovered that fine-tuning the key and value projections in cross-attention layers of the denoising network is the most effective way to achieve this. We set the input text condition to the optimized tokens. To avoid overfitting, we further append additional $L$ trainable tokens at the end of the text condition and apply prior preservation loss as in \cite{kumari_multi-concept_2022}. The training objective is as follows:
\begin{align}
    v_{[1:L]*}, \epsilon_{\theta*} = \mathop{\arg \min}\limits_{v_{[1:L]}, \epsilon_\theta} \: \mathbb{E}_{z_0 \in \mathcal{E}(I), y', \epsilon \sim \mathcal{N}(0, 1), t}&   \nonumber
    \\
    [ \| \epsilon - \epsilon_\theta(z_t, &t, c_\theta(y')) \|^2_2 ]
\end{align}
where $y' = [v_{1*},...,v_{N*}, v_{[1:L]}]$ and $v_{[1:L]}$ are the trainable appended tokens. Only key and value projections of cross-attention layers in $\epsilon_{\theta*}$ are finetuned.
}

\begin{algorithm}
\caption{Denoising process with layout control}
\label{algorithm}
\KwIn{The sequence of optimized text tokens for objects $y_*$, the initial image $I^*$,
a set of object masks $M$, optimization learning rate $\alpha_t$, a set of thresholds $\{Q_t\}$, the timestep to stop optimization and blending $t_{opt}$ and $t_{bld}$. }
\KwOut{An edited Image $I$}
\SetKw{KwGoTo}{go to}
Encode input image: $z_0^* = \mathcal{E}(I^*)$\;
Initialize with Gaussian noise: $z_T = \mathcal{N}(0,I)$\;

\For{$t = T,$ $...,$ $1$}{
    \tcp{Iterative optimization: }
    \If{$t \geq t_{opt}$}{
    Get cross attention: $A \gets \epsilon_{\theta*}(z_t, c_\theta(y_*), t)$\;
    \For{$k = 1,$ $...,$ $N$}{
        $A_{l,k} \gets A_l[:,:,k]$\;
        Calculate $\mathcal{L}_k$ with $M_k$ as in \refeqn{single_loss}\;
    }
    Calculate mean-max loss $\mathcal{L}$ as in \refeqn{loss}\;

    \If{$\mathcal{L} > 1 - Q_t$}{
    Update $z_t$ with: $z_t \gets z_t - \alpha_t \cdot \nabla_{z_t} \mathcal{L}$\;
    \If{ \textup{Reach maximum optimization steps} }{
    $\KwGoTo$ $22$\;
    }
    \Else{
    $\KwGoTo$ $5$\;
    }
    }
    }

\tcp{Background Blending:}
    \If{$t \geq t_{bld}$}{
        Add noise to original latent: $z_t^* = addnoise(z_0^*, t)$ \;
        Get mask of background: $M_{bg} = 1 - \sum_k M_k$ \;
        Blending:  $z_t \gets M_{bg} \odot z_t^* + (1 - M_{bg}) \odot z_t $ \;
    }
    $ z_{t-1} \gets denoising(z_t, c_\theta(y_*), t)$ \;
}
Decode the edited image: $I = \mathcal{D}(z_0)$
\end{algorithm}

\subsection{Training-Free Layout Editing }
\label{sec:iter_optim}
\begin{figure}[htb]
    \centering
    \includegraphics[width=1.0\linewidth]{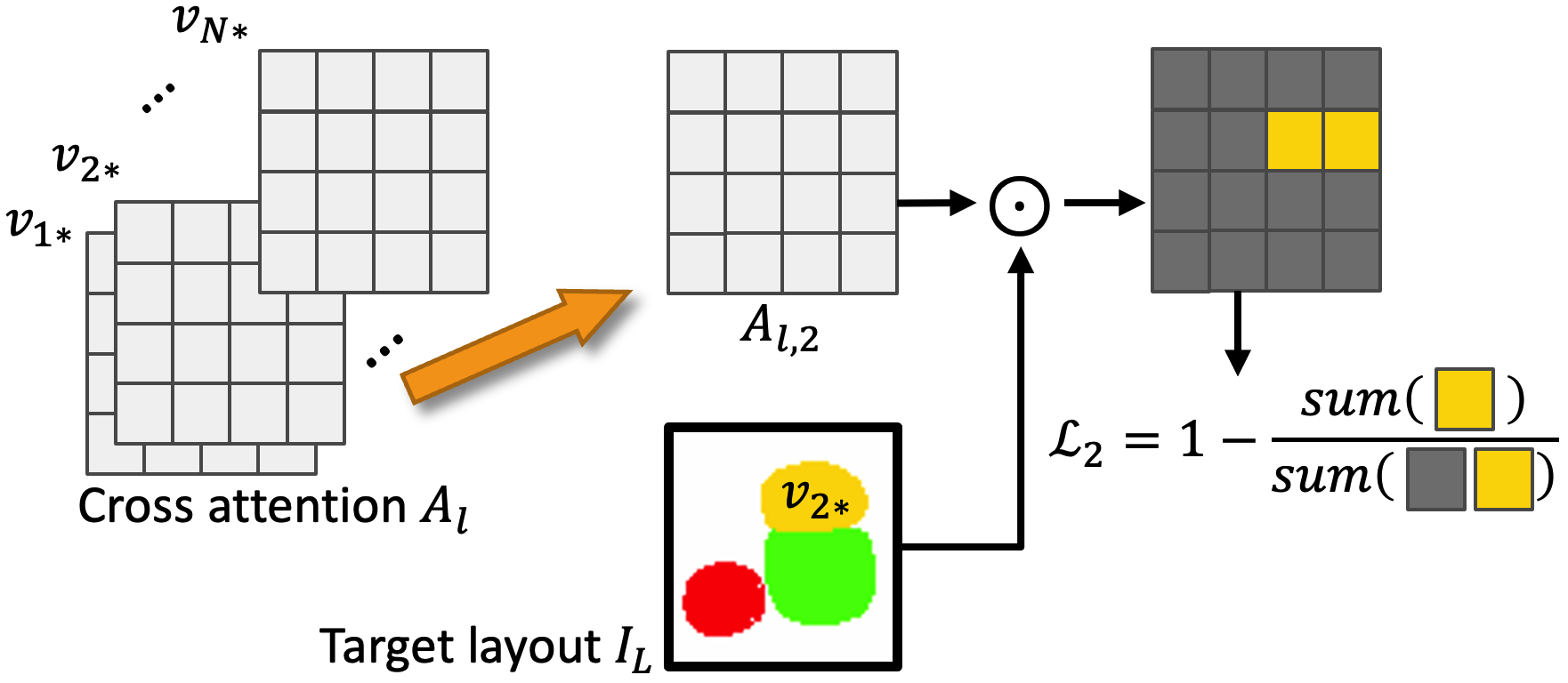}
    \caption{Region loss calculated from cross attention. It encourages higher cross-attention between the object's text embedding and its corresponding region than with any other region.}
\label{fig:loss}
\end{figure}

\zt{
With the learned concepts of multiple objects $v_{1*},...,v_{N*}$ and the fine-tuned model $\epsilon_{\theta*}$, we rearrange the positions of the objects to edit the layout. A straightforward way to control the layout is to add a new layout condition to a stable diffusion model, as in~\cite{avrahami_spatext_2023,zhang_adding_2023,li_gligen_2023}. However, this approach requires further fine-tuning with a additional dataset. Instead, we propose a training-free method to control the layout to avoid dataset collection. 

As discovered in P2P~\cite{hertz_prompt--prompt_2022}, the cross-attention in the denoising network of text-to-image diffusion models can reflect the positions of each generated object specified by the corresponding text token, which is calculated from:
\begin{equation}
    A_l = \sigma ( Q_l(z_t^l) K_l(y)^T )
\end{equation}
where $A_l$ is the cross-attention at layer $l$ of the denoising network, $Q_l, K_l$ are the query and key projections, $\sigma$ is the softmax operation along the dimension of text embedding $y$, and $z_t^l$ is the intermediate feature of the image latent. The calculated attention $A_l$ has the size of $h_l \times w_l \times d$, where $h_l$ and $w_l$ is the spatial dimension of the feature $z_t^l$ and $d$ is the length of input text tokens. More specifically, 
for each text token, we could get an attention map of size $h_l \times w_l$ which reflects the relevance to its concept. For example, in the attention map with the text "cat", the positions within the area containing the cat should have larger values than other positions.
}
\zt{
Therefore, we could optimize $z_t$ towards the target that the desired area of the object has large values. Previous study~\cite{hertz_prompt--prompt_2022} has shown that the layers of resolution $16\times16$ contain the most meaningful semantic information. Therefore, we choose $l$ to be the layers with $h_l = w_l = 16$. As shown in \reffig{loss}, for each layer $l$, each channel of the cross attention $A_l$ represents the spatial relevance to the corresponding text token. For example, if we want to optimize the position of the object represented by $v_{2*}$, we can extract the corresponding channel $A_{l,2}$ and multiply it with the target region mask of $v_{2*}$ (i.e., the yellow mask). Finally, the region loss can be calculated with the summation of the values within the mask (yellow region) and of all positions (black and yellow regions):
\begin{align}
    \mathcal{L}_k &=  1 - \frac{\sum_i (M_k^* \odot \sum_{l}A_{l,k})}{\sum_i \sum_{l}A_{l,k}}
    \label{eqn:single_loss}
    \\
    \mathcal{L} &= \frac{1}{N} \sum_{k=1}^N \mathcal{L}_k + \max (\mathcal{L}_1,...,\mathcal{L}_N)
    \label{eqn:loss}
\end{align}
where $\mathcal{L}_k$ is the loss for the $k$-th object with target position specified by $M_k^*$, $A_{l,k}$ is the cross-attention map with the optimized token $v_{k*}$ (from \refeqn{opt_text}) at layer $l$, and $i$ is the spatial position in $A_{l,k}$. We use the mean value together with the maximum value of $\mathcal{L}_k$, so that the model can control the positions of each object and simultaneously focus on the object with a large loss.
}

\zt{
We optimize the latent $z_t$ with the loss $\mathcal{L}$ in \refeqn{loss} only at large timesteps, i.e., $t >= 0.5$, which is enough to fix the layout of the generated image. We apply iterative optimization for $t=1.0,0.8,0.6$ with maximum steps and early stopping. For other timesteps, we only update $z_t$ for one single step. Although the model has memorized the background during the process mentioned in \refsec{mask_text_inv}, it still introduces distortions to the background during the optimization for layout control. To preserve the original background, we start to blend with the original input image at the area without objects as in~\cite{avrahami2022blended}, for timesteps $t >= 0.7$. The detailed algorithm for our layout control method during the denoising process is shown in Algo.~\ref{algorithm}. 
}

\subsection{Implementation Details}



We optimize each token for different objects using masked textual inversion for 200 steps with a batch size of 4. The learning rate is set to 0.002, and it takes approximately 40 seconds for each token on a single Nvidia V100 GPU. Next, we fine-tune the model using all the optimized tokens concatenated with 1-3 additional rare tokens for 800 steps (for 2 or 3 objects) or 1200 steps (for 4 objects), with a batch size of 4. The learning rate is set to 0.0002, and it takes around 3-4 minutes on an Nvidia V100 GPU. To sample images, we use DDIM sampling~\cite{song_denoising_2022} with 50 steps. For layout control, we optimize $z_t$ with a learning rate that decreases from 20 to 15 for $t$ values from 1.0 to 0.5. We apply iterative optimization for 40 iterations with early stopping thresholds of 0.4, 0.3, and 0.2 when $t$ is 1.0, 0.8, and 0.6, respectively. It takes around 37 seconds to generate an image with a new layout.



\section{Experiments}
To evaluate our proposed method, we first compare it qualitatively and quantitatively with several baselines, and then conduct a user study to compare them perceptually. Additionally, we conduct several ablation studies to validate the effectiveness of each important component of our method and demonstrate its application in continuous layout editing, which was not feasible with previous methods.

\subsection{Baselines}
\label{sec:comparison}

\begin{figure*}[tbp]
    \centering
    \includegraphics[width=1.0\linewidth]{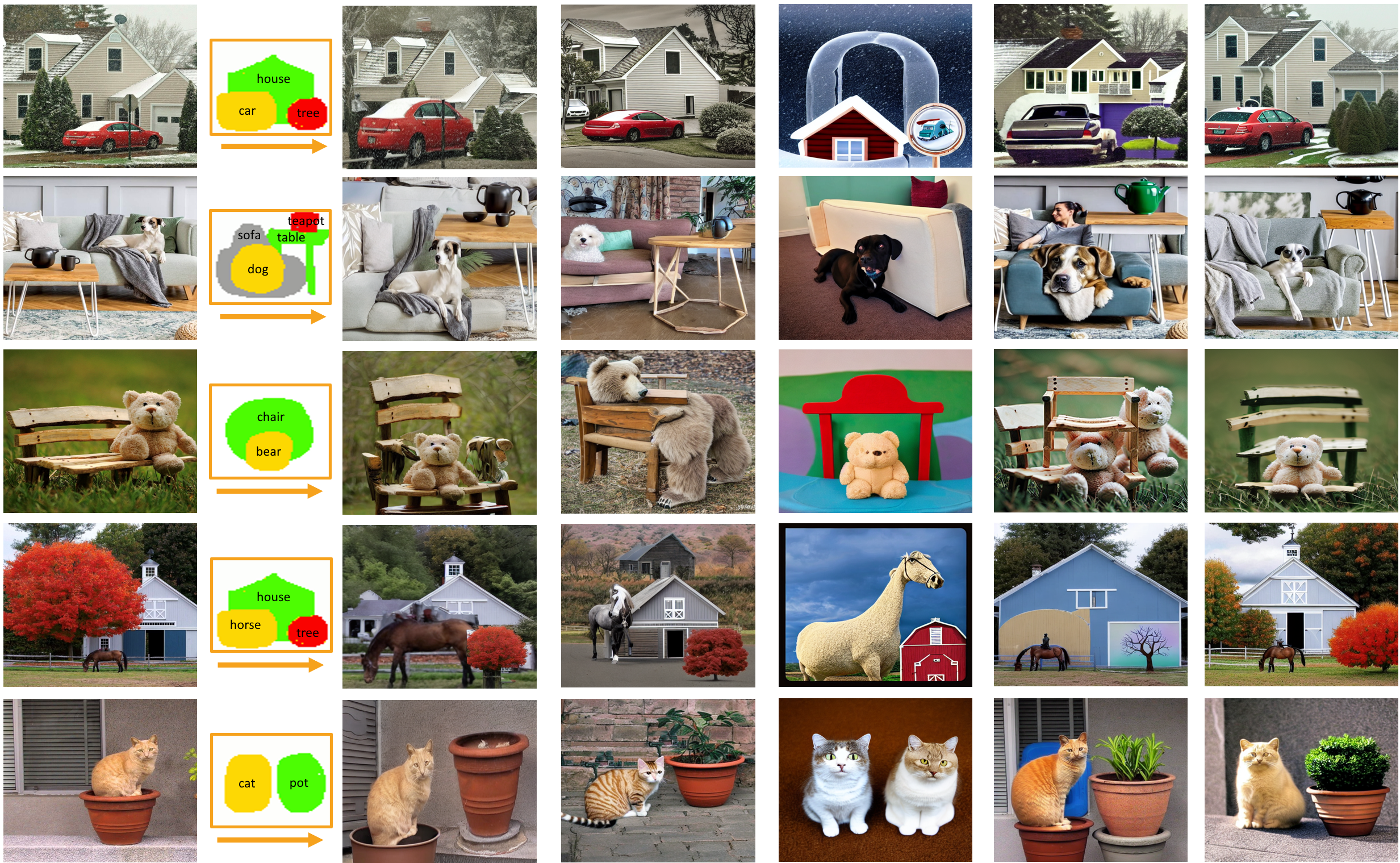}
    \parbox[t]{1.0\linewidth}{\relax
           \textit{ \hspace{25px} (a) \hspace{46px} (b) \hspace{46px} 
           (c) \hspace{63px} (d) \hspace{63px} (e) \hspace{63px} (f) \hspace{63px} (g) }
           }
    \caption{Qualitative comparison with other baseline methods. From left to right: (a) Input images; (b) Target layout; (c) Image-level manipulation; (d) Latent-level manipulation; (e) GLIGEN + textual inversion; (f) MultiDiffusion + Dreambooth; (g) Ours}
\label{fig:compare}
\end{figure*}

Since there are no existing methods that perform the same task as ours, which is to edit the layout of existing images, we compare our method with four baselines that are designed and modified from existing methods: image-level manipulation, noised latent-level manipulation, and two variations of combining existing layout control methods with textual inversion, which include GLIGEN~\cite{li_gligen_2023} with textual inversion~\cite{gal_image_2022} and MultiDiffusion~\cite{bar-tal_multidiffusion_2023} with Dreambooth~\cite{ruiz_dreambooth_2023}.

\textbf{Image-level manipulation.} We first crop and paste the objects from the original input image onto a blank image at the positions specified in the target layout. Since the target layout may specify objects with different widths and heights, we scale the cropped patches to match the desired size. Next, we use stable diffusion inpainting to fill in the blank areas.

\textbf{Latent-level manipulation.} Instead of cropping and pasting on the image-level, we perform a similar process on the noised latent from DDIM inversion~\cite{song_denoising_2022}. We initialize a random noise as the "canvas" and prepare a "source image" by adding noise to the original input image until $t=0.7$ using DDIM inversion. Then, we copy objects in the "source image" and paste them onto the "canvas" following the target layout without scaling, as resizing in latent space can lead to distorted results. Finally, we use the DDIM scheduler to denoise the "canvas" and obtain the result..

\textbf{GLIGEN with textual inversion.} Although GLIGEN~\cite{li_gligen_2023} can perform layout control, it can only generate new images with target layouts. Therefore, to adapt it to our task, we need to add an additional step of textual inversion~\cite{gal_image_2022} to learn the appearances of the objects in existing images. After textual inversion, we use the learned text tokens to sample the image, where the layout is controlled with GLIGEN.

 \textbf{MultiDiffusion with Dreambooth.} MultiDiffusion~\cite{bar-tal_multidiffusion_2023} is a training-free method for layout control, but it cannot be directly used for existing image layout editing. Therefore, before using it, we adopt Dreambooth~\cite{ruiz_dreambooth_2023} to learn the concepts of the objects in the input image, and convert them into multiple text tokens. Then, we provide the prompt with the learned text tokens and corresponding masks to MultiDiffusion to perform our task of layout editing.


\subsection{Qualitative Comparison}
\reffig{compare} illustrates the qualitative comparisons between our method and four baselines. The image-level manipulation method (column c) produces less realistic results as it only copies and resizes objects to the target positions without natural editing. For example, in row 4, the size of the horse is significantly larger than that of the tree, violating the proper order of sizes. Moreover, noticeable artifacts appear around objects due to imperfect cropping and inpainting, as evidenced by the distorted chair armrests in row 3 and the artifacts on the horse's back in row 4. The noise-level manipulation (column d) can maintain the basic layout but cannot retain the visual features of original objects since DDIM inversion cannot reconstruct the input image, resulting in the loss of the "dog" appearance in row 2. GLIGEN with textual inversion (column e) also fails to retain the visual properties of objects since the textual inversion on the whole image cannot disentangle the concepts of multiple objects within a single image. In row 5, for instance, the concept of the "pot" is not correctly learned. Additionally, GLIGEN is pre-trained with a dataset to perform layout control, which cannot be perfectly adapted to learned concepts from a particular image, leading to some misalignment with the specified layout, as shown in row 4. MultiDiffusion with Dreambooth (column f) has better layout alignment than GLIGEN with textual inversion because MultiDiffusion is a training-free method that can better cooperate with learned concepts. However, Dreambooth is designed for learning concepts of single objects from multiple images and still cannot disentangle multiple objects present in a single image, causing significant changes in object characteristics, as shown by the car in row 1 and the dog in row 2. Furthermore, it introduces artifacts around the boundaries of objects because MultiDiffusion denoises each sub-region with different diffusion processes. In row 4, for instance, the regions of the "horse" and the "tree" are separately generated, resulting in artifacts around the objects and discontinuity between objects, such as the yellow background around the "horse" and the cyan background around the "tree". Our method (column g) achieves the best results by effectively controlling the layout of objects while retaining the visual features of the input images. It also produces the highest quality and most harmonious images among all four baselines.
\subsection{Quantitative Comparison}
\begin{table}[t]
\label{tab:metric}
\begin{minipage}{\columnwidth}
\begin{center}
\caption{Quantitative comparison with other baseline methods. Our method achieves the best performance on both metrics: visual similarity to the input image and alignment with the specified layout.}
\begin{tabular}{ccc}
  \toprule
    {} &  Visual & Layout  \\ 
    {} &  similarity $\uparrow$ & alignment $\uparrow$ \\ \toprule
    Image-level manipulation & 0.57 & 0.0068 \\ \midrule
    Latent-level manipulation & 0.41 & 0.0071 \\ \midrule
    GLIGEN with textual inversion & 0.34 & 0.0027 \\ \midrule
    MultiDiffusion with Dreambooth & 0.53 & 0.0047 \\ \midrule
    Ours & \textbf{0.61} & \textbf{0.0099} \\
  \bottomrule
\end{tabular}
\end{center}
\end{minipage}
\end{table}



We conduct a quantitative study to further evaluate the preservation of visual properties and layout alignment of our method and compare it with baselines. To measure the visual similarity between the edited result and the input image, we calculate one minus the CLIP distance between the input image and edited image. A higher score indicates that the objects in the edited image have a more similar visual appearance to the input image. For layout alignment, we calculate the CLIP distance between the image and text prompt. Specifically, for an object in the layout map, we erase the corresponding region in the edited image and fill it with black color. We then calculate the change in CLIP distance between the image (before and after erasing) and the text token corresponding to the object. If the CLIP distance drops dramatically after erasing, it indicates that the object is placed in the correct position after layout editing. This process is repeated for each object, and an average score is calculated. The experiment shows that our method has the best performance in both metrics: the least image-to-image distance (0.61) and the largest change in image-to-text distance (0.0099).

\begin{figure}[htb]
    \centering
    \includegraphics[width=1.0\linewidth]{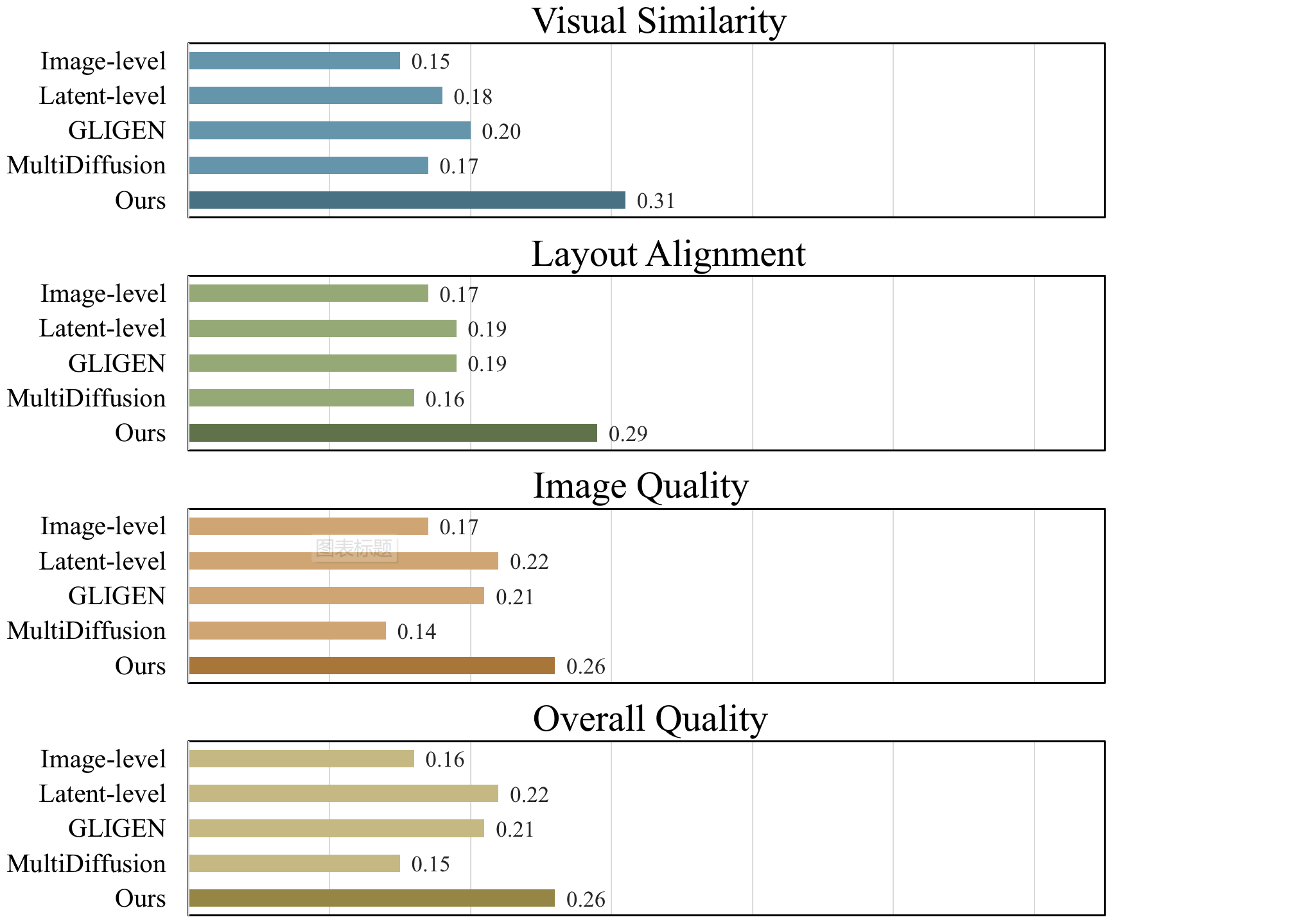}
    \caption{Results of the user study on visual similarity, layout alignment, image quality, and overall quality, respectively.}
\label{fig:user_study}
\end{figure}

\subsection{User Study}

In this part, we perform a user study to verify the effectiveness and quality of our method. We compare our method with four baselines mentioned in \refsec{comparison}, \ie, Image-level manipulation, Latent-level manipulation, GLIGEN with textual inversion, and MultiDiffusion with Dreambooth.

The user study consists of 16 questions. For each of the questions, we first show the original input image and the target layout to the user. Then we show the five images edited by four compared methods and our method in random order. Finally, the user is asked to make four selections regarding four different factors:
\begin{itemize}
    \item Visual similarity: to choose the image whose generated objects have the highest similarity to the objects in the original input image.
    \item Layout alignment: to select the image whose layout is best aligned with the target layout.
    \item Image quality: to select the image with the highest quality and photorealism.
    \item Overall quality: to choose the best result considering all the three factors above.
\end{itemize}
Among the 16 questions, we also set a validation question where the 5 edited images consist of a ground-truth image with four unrelated images. The user has to make more than 3 correct selections out of 4, to be considered as a valid questionnaire.

We finally collected 42 questionnaires, among which 31 are valid by passing the validation questions. Among the 31 valid participants, 5 users are below 20 years old, 17 range from 20 to 30 years of age, 6 are between 30 and 40 years old, and 3 are above 40 years old. The result of the user study are shown in \reffig{user_study}, and we find that our method outperforms other methods in all the four factors with a preferred rate of $31\%$ in visual similarity, $29\%$ in layout alignment, $26\%$ in image quality, and $26\%$ in overall quality.

\begin{figure}[htb]
    \centering
    \includegraphics[width=1.0\linewidth]{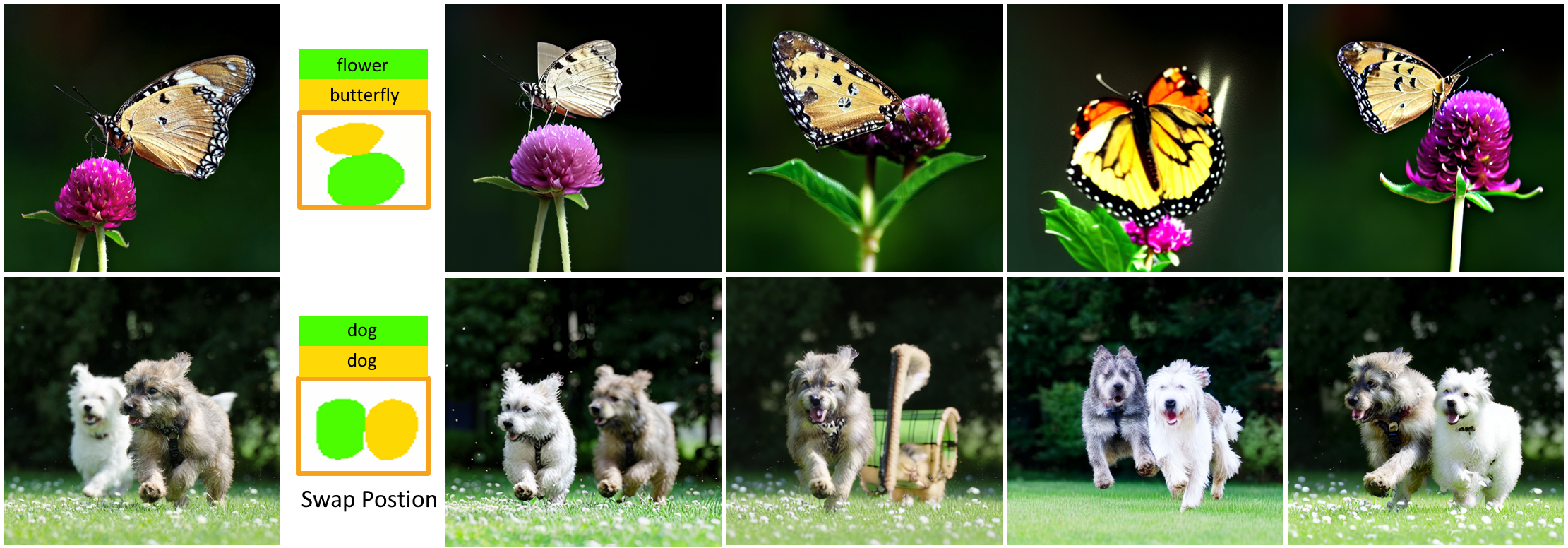}
    \parbox[t]{1.0\linewidth}{\relax
           \textit{ \hspace{11px} (a) \hspace{19px} (b) \hspace{19px} 
           (c) \hspace{29px} (d) \hspace{29px} (e) \hspace{29px} (f) }
           }
    \caption{Ablation study on different textual inversion methods. From left to right: (a) Input images; (b) Target layouts; (c) Inversion with Dreambooth~\cite{ruiz_dreambooth_2023}; (d) Textual inversion~\cite{gal_image_2022} + finetune~\cite{kumari_multi-concept_2022}; (e) Masked textual inversion w/o finetune; (f) Our full inversion method with masked textual inversion and finetune. }
\label{fig:abl_mask}
\end{figure}

\subsection{Ablation Study}

\subsubsection{Inversion Methods}
\zy{In this part, we compare the inversion method with and without a mask. We selected Dreambooth~\cite{ruiz_dreambooth_2023} and textual inversion~\cite{gal_image_2022} + finetune as representatives of the methods without a mask. For Dreambooth, we jointly trained the model on multiple concepts with a single image and prompt pair, while the textual inversion + finetune method first runs the textual inversion to update each token on the entire image and then uses the updated tokens to fine-tune the cross-attention layers of the model. 
The results show that if a single image contains multiple objects, the methods without a mask, including both Dreambooth and textual inversion, cannot precisely learn and disentangle the concepts of different objects, which may result in the loss of visual properties. For example, in \reffig{abl_mask}, row 1, the shape and color of the flower change significantly in columns (c) and (d). Another issue with these methods without a mask is learning incorrect objects. For instance, in \reffig{abl_mask}, row 2, columns (c) and (d), the information about the white and brown dogs is not encoded into two separate tokens. Therefore, both methods cannot correctly swap the position of the two dogs. The textual inversion + finetune method (d) even generates a trolley-like object instead of a dog.

In our method, we optimize the text tokens using masked textual inversion followed by model finetuning, as described in \refsec{mask_text_inv}. By applying the mask, the information of different objects in the image is correctly disentangled. In \reffig{abl_mask}, row 2, columns (e) and (f), the two dogs successfully swap positions. Moreover, adding finetuning after our masked textual inversion can further preserve visual details, including colors and textures.}

\begin{figure}[htb]
    \centering
    \includegraphics[width=1.0\linewidth]{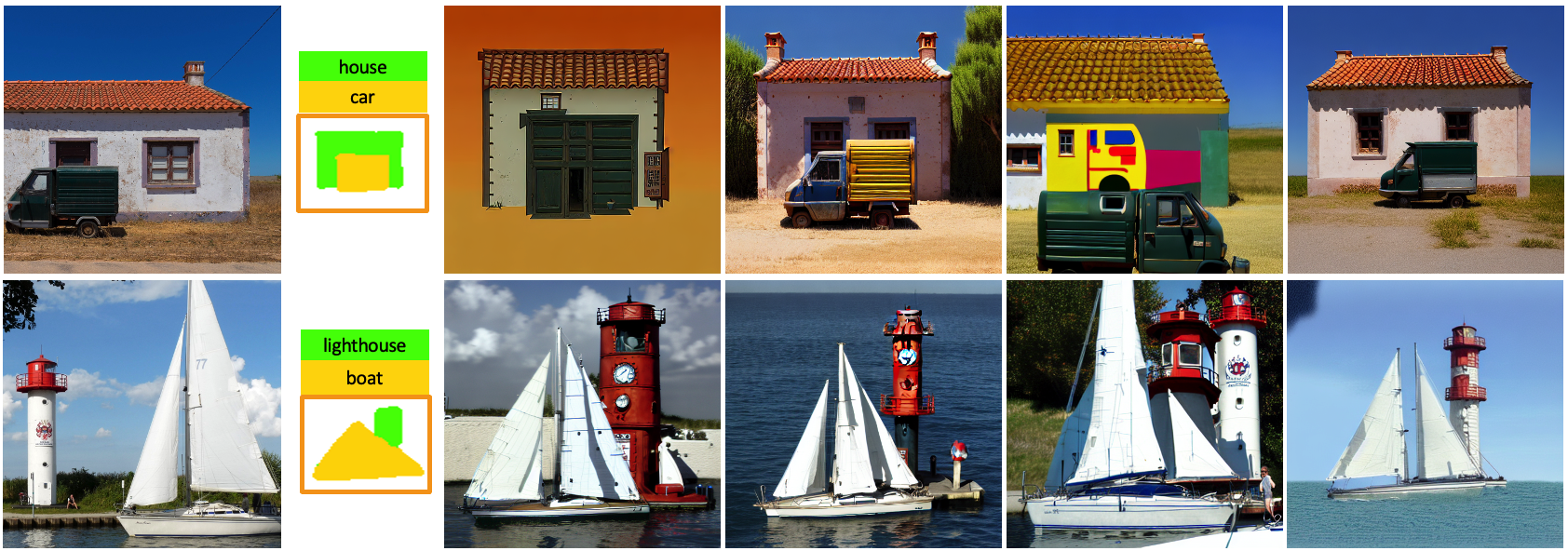}
    \parbox[t]{1.0\linewidth}{\relax
           \textit{ \hspace{11px} (a) \hspace{19px} (b) \hspace{19px} 
           (c) \hspace{29px} (d) \hspace{29px} (e) \hspace{29px} (f) }
           }
    \caption{Ablation study on differen layout control methods. From left to right: (a) Input images; (b) Target layouts; (c) Our inversion + ControlNet~\cite{zhang_adding_2023}; (d) Our inversion + GLIGEN~\cite{li_gligen_2023}; (e) Our inversion + MultiDiffusion~\cite{bar-tal_multidiffusion_2023}; (f) Our full methods. }
\label{fig:abl_model}
\end{figure}

\begin{figure}[htb]
    \centering
    \includegraphics[width=1.0\linewidth]{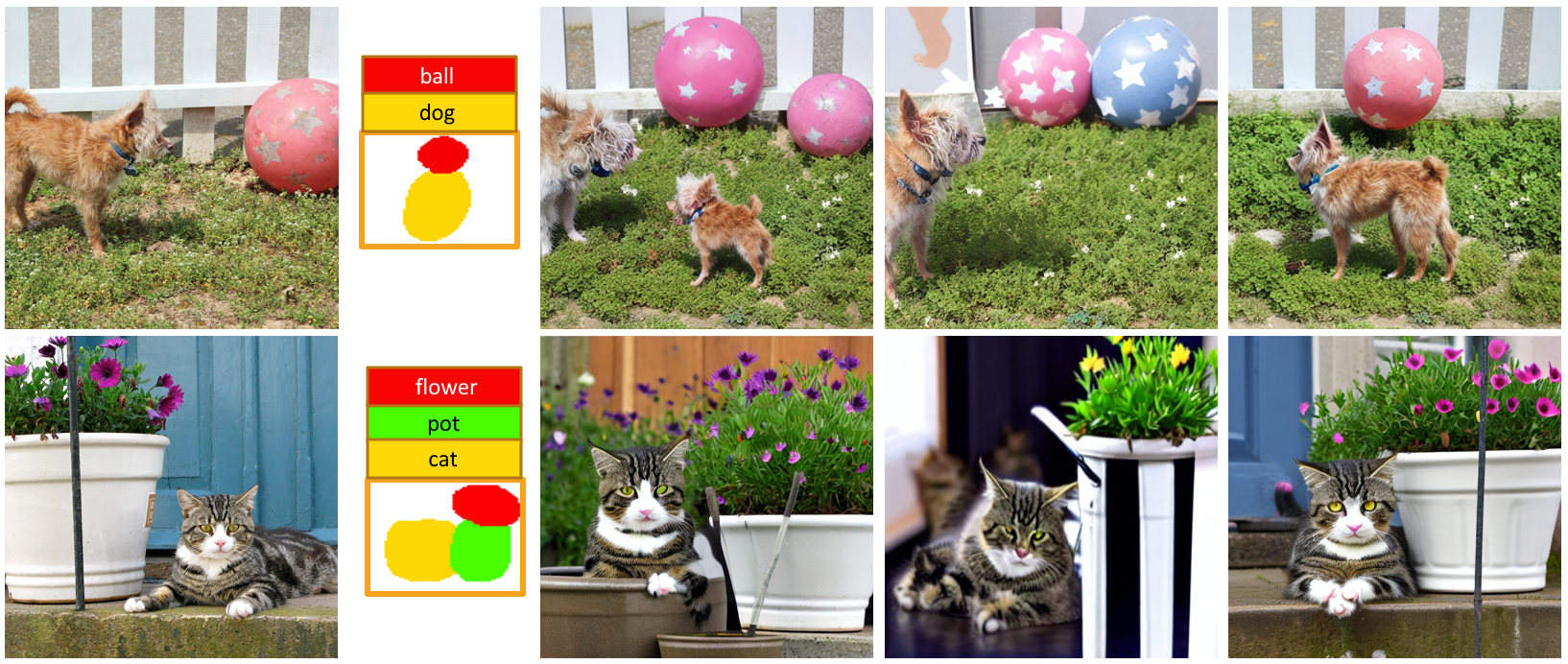}
    \parbox[t]{1.0\linewidth}{\relax
           \hspace{1px} Input images \hspace{35px} 
           Mean loss \hspace{14px} Max loss \hspace{10px} \parbox[t]{0px}{Mean+max (ours)}
           }
    \caption{Ablation study on optimization loss of layout control.}
\label{fig:abl_loss}
\end{figure}

\begin{figure}[htb]
    \centering
    \includegraphics[width=1.0\linewidth]{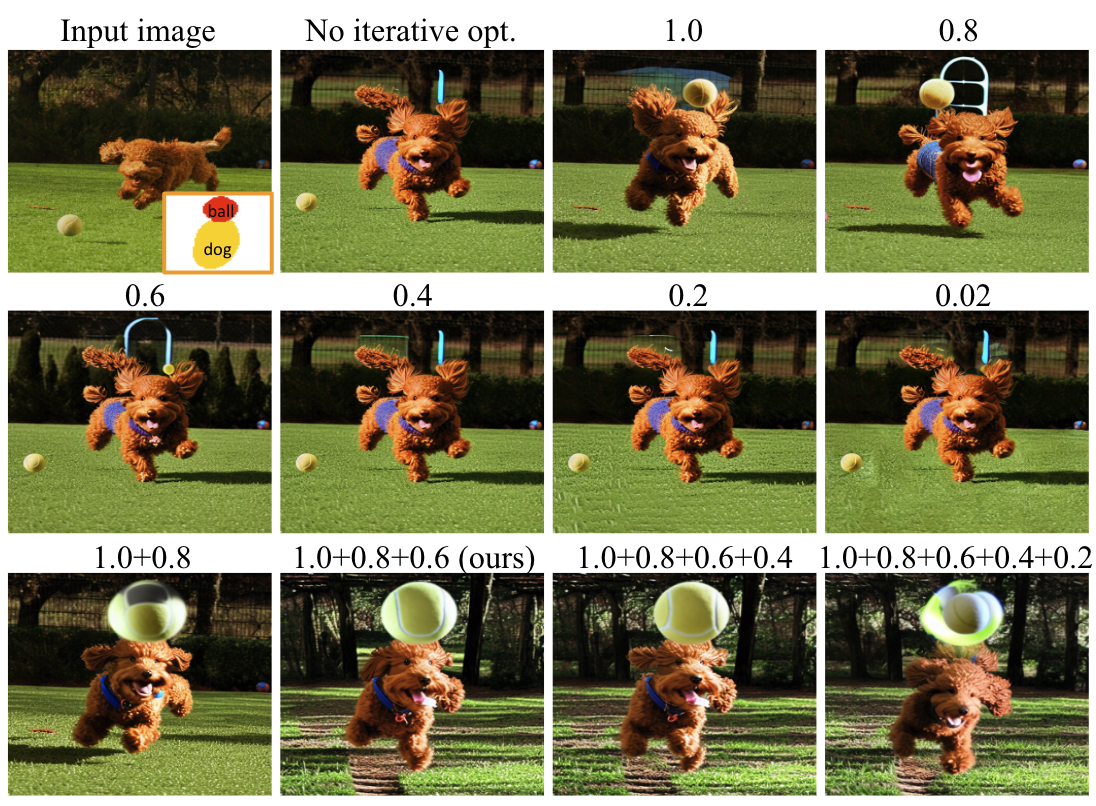}
    \caption{Ablation study on iterative optimization. The number above the image indicates the denoising step on which iterative optimization is applied. If more than one number is labeled, iterative optimization is applied to multiple denoising steps.}
\label{fig:abl_iter}
\end{figure}

\begin{figure}[htb]
    \centering
    \includegraphics[width=1.0\linewidth]{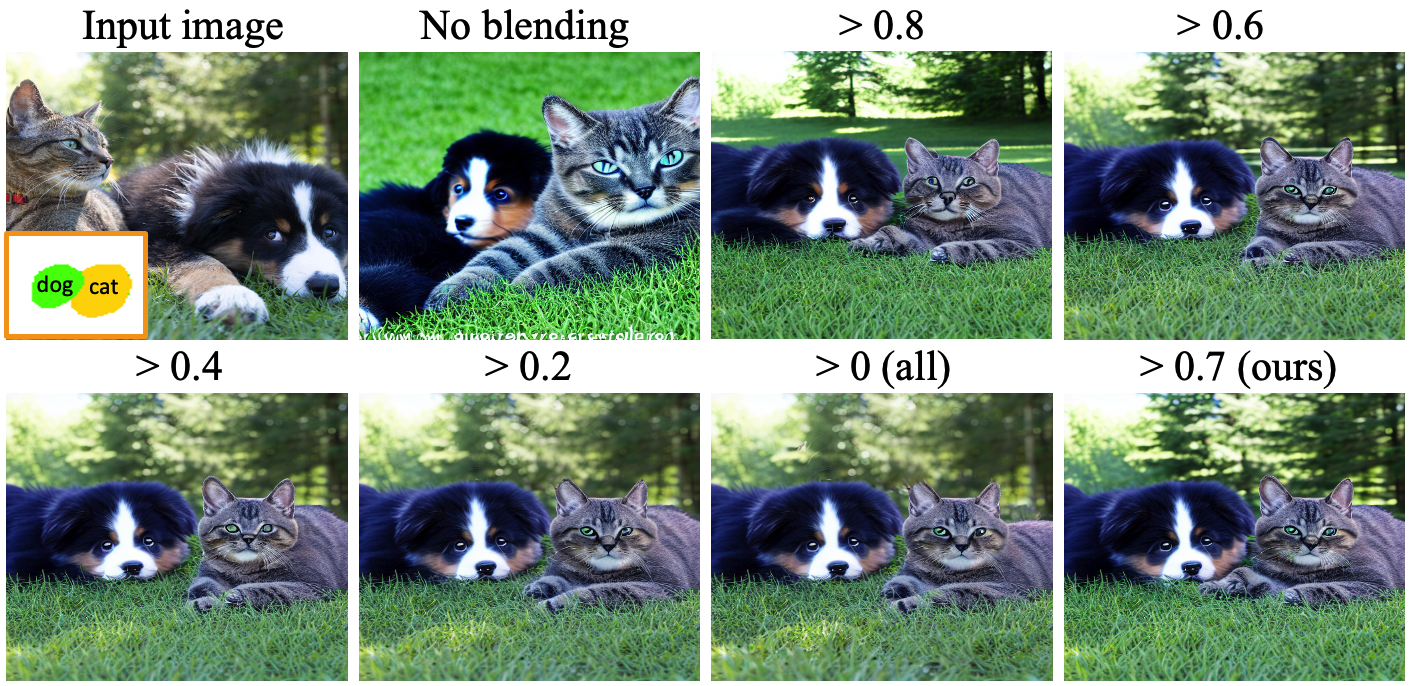}
    \caption{Ablation study on blending steps. The number above the image indicates the number of steps where blending is applied. }
\label{fig:abl_blend}
\end{figure}

\subsubsection{Layout Control Methods}
After verifying the effectiveness of our masked textual inversion, we compared our layout control method with three other methods: ControlNet~\cite{zhang_adding_2023}, GLIGEN~\cite{li_gligen_2023}, and MultiDiffusion~\cite{bar-tal_multidiffusion_2023}. For the training-based methods ControlNet and GLIGEN, the generated objects fail to keep some visual features in the input images, such as the "car" in row 1 and the "lighthouse" in row 2 of \reffig{abl_model}. This is because those models are pretrained with datasets that may not be well-adapted to our learned concepts, especially considering that our finetuning after masked textual inversion may further change the parameters of the cross-attention layers, which can affect the ability of the pretrained model. As for the training-free method MultiDiffusion, it can better incorporate learned concepts, but it tends to yield some artifacts around the objects as it denoises each sub-region separately and fails to fuse them smoothly. For example, there is a yellow painting on the house in \reffig{abl_model}, row 1, column (e). Our iterative layout control method solves the aforementioned problems and generates the best results in (f)

\begin{figure*}[tbp]
    \centering
    \includegraphics[width=1.0\textwidth]{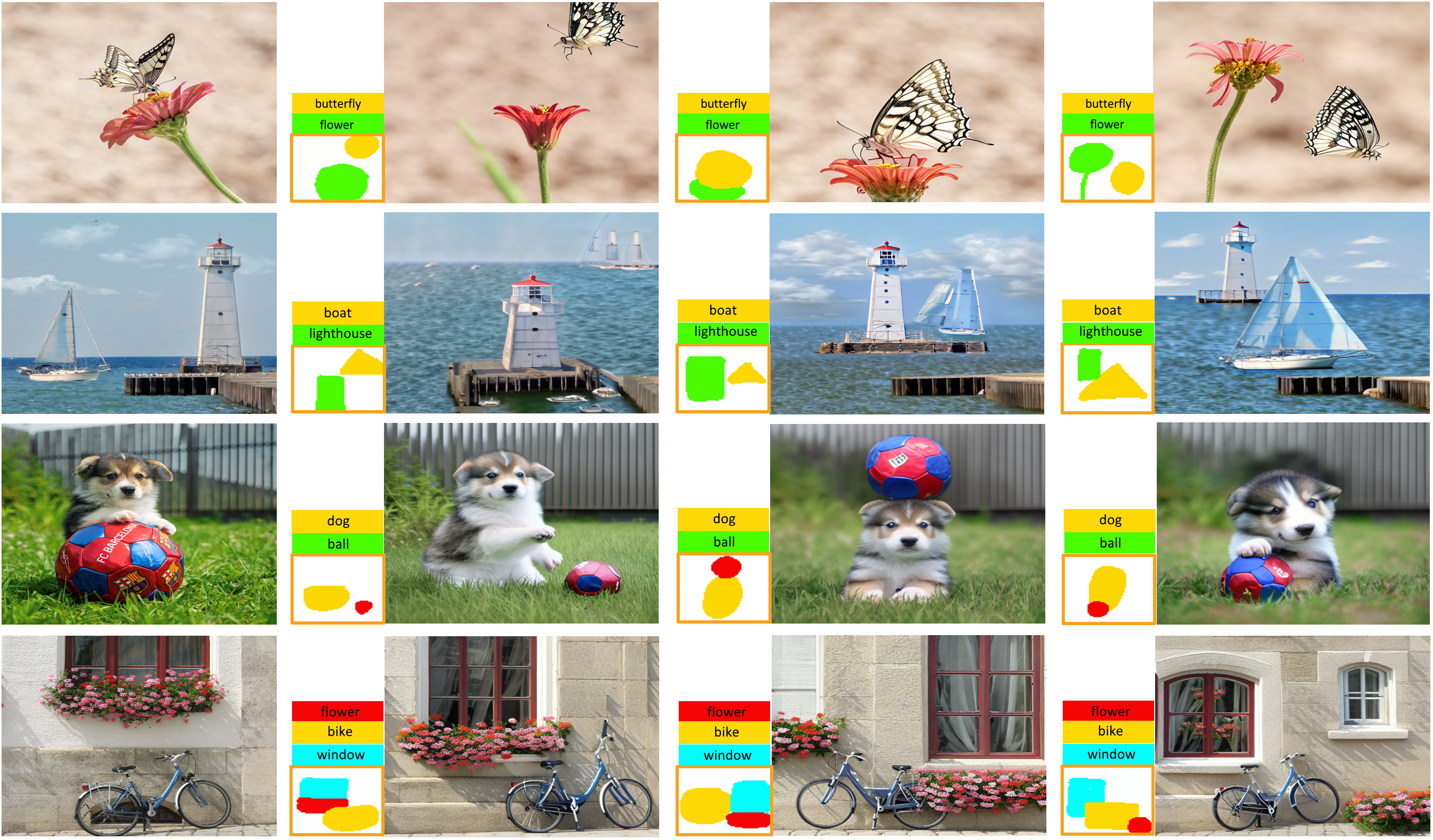}
    \caption{Continuous layout editing with different target layouts. }
\label{fig:application}
\end{figure*}

\subsubsection{Optimization Loss}
\zt{
In \refeqn{loss}, we mention that both the mean and the maximum values of ${ \mathcal{L}_k }$ are calculated for the optimization loss. As shown in \reffig{abl_loss}, if only the mean loss is applied, the controls against each object are equal, and no additional effort can be put on the difficult one. Therefore, the positions of some objects may be insufficiently controlled. For example, part of the "dog" still stays at the original position in \reffig{abl_loss}, row 1, and the relative position of the "dog" and the "pot" is not correctly arranged in row 2.

On the other hand, if only the max loss is applied, the model may focus too much on single object, but ignore others. For example, the position of the "dog" is not modified in \reffig{abl_loss}, row 1, and the "pot" is stretched and distorted to fit the target position in row 2. Therefore, we finally apply both the mean and max loss, which can balance the layout control of each object and simultaneously focus more on objects that are difficult to control.
}

\subsubsection{Iterative Optimization}
\zt{
As described in Section \ref{sec:iter_optim}, we only apply iterative optimization at certain time steps. In this part, we performed experiments to verify the effect of iterative optimization and determine which time steps to apply it. As shown in Figure \ref{fig:abl_iter}, when iterative optimization is not applied, the objects are not well aligned with the target layout.

We also found that the determination of the layout control happens at the time steps closer to the noise. If iterative optimization is applied at large time steps (e.g., $t = 1.0$, $0.8$, or $0.6$), it is more effective for layout control, as a ball appears in the upper part of the image. Conversely, with small time steps (e.g., $t = 0.4$, $0.2$, or $0.02$), the layout has little change. This implies that the layout of the objects is nearly fixed at large $t$ and can hardly be modified when $t$ is small. Therefore, we implement iterative optimization at relatively larger time steps, i.e., $t = 1.0 + 0.8 + 0.6$.

As shown in the image, our choice can generate a ball with the desirable size. However, fewer iterative optimization time points (e.g., $t = 1.0 + 0.8$) may not be sufficient for generating the ball completely. Conversely, too many iterative optimization time points (e.g., $t = 1.0 + 0.8 + 0.6 + 0.4$, $t = 1.0 + 0.8 + 0.6 + 0.4 + 0.2$) have little effect on the final layout but require longer optimization time and may introduce artifacts. Thus, we choose $t = 1.0 + 0.8 + 0.6$ for a balance between quality and speed.
}


\subsubsection{Blending}
\zt{As described in Section \ref{sec:iter_optim}, we blend the edited image with the original input at the region of the background. In this part, we perform experiments to evaluate the effect of the different number of blending steps. As shown in Figure \ref{fig:abl_blend}, the background will be completely changed if no blending is applied because the optimization process for layout control will have a large influence on the background. Blending when $t$ is large (e.g., $t>0.8$) has the most substantial effect on the background, and the effect of blending starts to converge for more steps when $t>0.6$. Therefore, we choose to apply background blending when $t>0.7$.}
\subsection{Results of Continuous Editing}

Our method is capable of rearranging the positions of objects in an input image to fit a target layout without altering its visual properties. This unique capability enables our method to perform continuous layout editing of single input images, which was not possible with previous methods. Figure \ref{fig:application} and Figure \ref{fig:teaser} demonstrate some examples of our method's effectiveness. For instance, in row 3 of Figure \ref{fig:application}, we show how our method can change the positions of the dog and ball to fit three different layouts, creating natural interactions between them for each layout. Moreover, our method can handle images with multiple objects. We demonstrate this by showing how it can edit the positions of three objects in Figure \ref{fig:application}, rows 4, and four objects in Figure \ref{fig:teaser}, row 2, to align with different input layouts. These examples highlight the flexibility of our method to handle varying numbers of objects of different categories and sizes.

\section{Conclusions \& Limitations}

We present the first framework that supports continuous layout editing of single images, generating high-quality results by rearranging the positions of objects in the input image to fit a user-specified layout while preserving their visual properties. A key component enabling us to learn objects from a single image is Masked Textual Inversion, which disentangles multiple concepts into different tokens. With learned objects, we propose a training-free iterative optimization method for layout control. We demonstrate that our framework outperforms other baselines, including image-level manipulation, latent-level manipulation, and combinations of existing learning and layout control methods.

\begin{figure}[htb]
    \centering
    \includegraphics[width=1.0\linewidth]{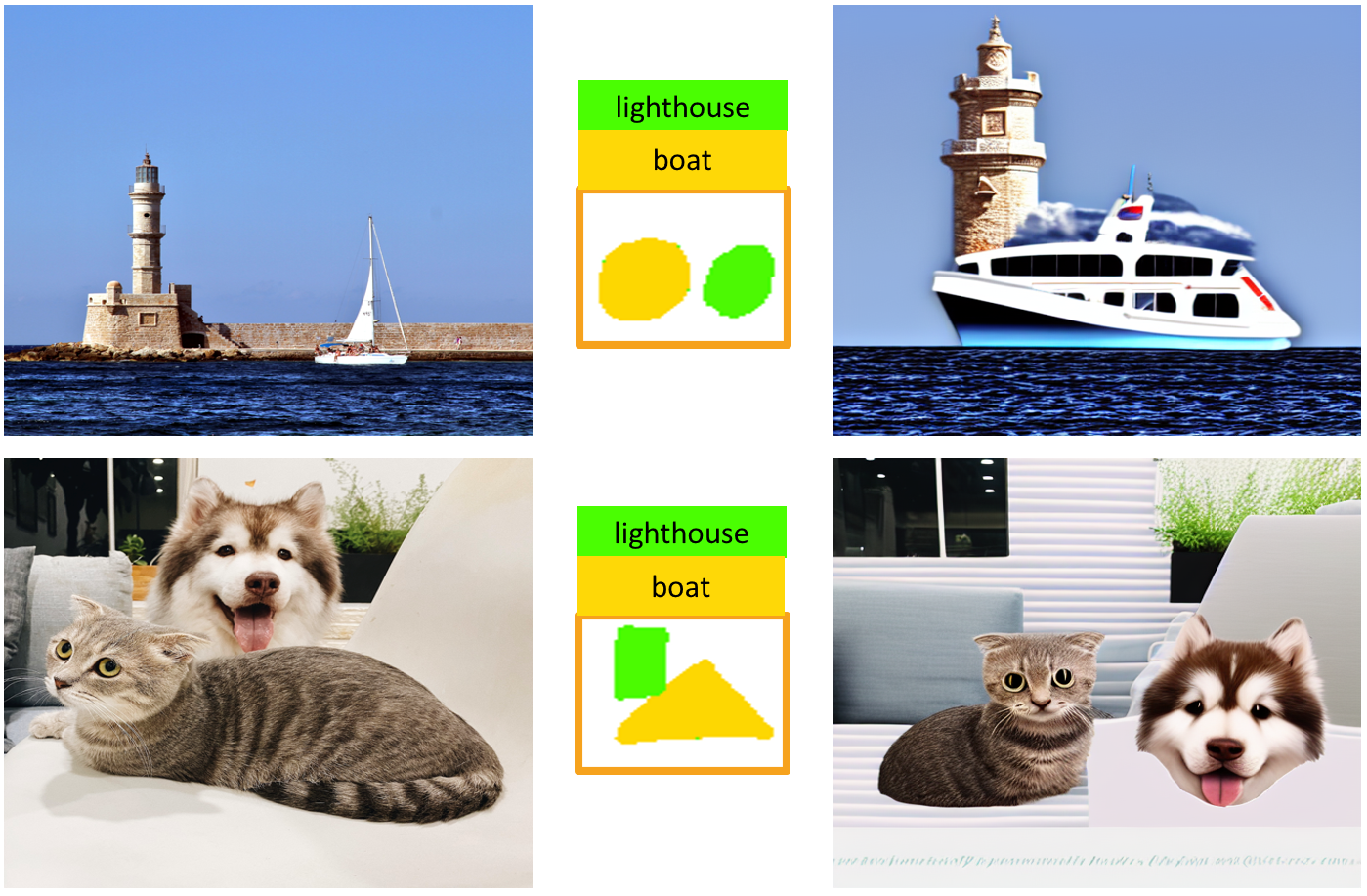}
    \parbox[t]{1.0\linewidth}{\relax
           \hspace{20px} Input images \hspace{94px} Edited images
           }
    \caption{Failure cases: In the first row, our method fails to maintain the visual features of objects when there is a significant size difference between the input and edited images. In the second row, our method fails to generate the full body of objects when they suffer from large occlusions in the original image.}
\label{fig:limitation}
\end{figure}

However, our method still encounters some limitations. One limitation is that it may fail to preserve the visual details of an object if the size of the object in the initial image and edited image varies significantly, as shown by the sailboat in Figure \ref{fig:limitation}, row 1. Another limitation is that it may have difficulty recovering the full body of an object if the object in the input image is heavily occluded, as shown by the dog in Figure \ref{fig:limitation}, row 2. We believe that these problems are caused by the limited information that can be inferred from a single image by object concept learning. To mitigate these limitations, we could augment the input images to different sizes and angles and even inpaint missing parts of the object caused by occlusion before applying our masked textual inversion. Furthermore, our layout editing is not in real-time due to the iterative sampling nature of the diffusion model. Future research directions include exploring methods to accelerate the process and supporting more applications of layout editing.




\bibliographystyle{ACM-Reference-Format}
\bibliography{reference}


\begin{thebibliography}{18}


\ifx \showCODEN    \undefined \def \showCODEN     #1{\unskip}     \fi
\ifx \showDOI      \undefined \def \showDOI       #1{#1}\fi
\ifx \showISBNx    \undefined \def \showISBNx     #1{\unskip}     \fi
\ifx \showISBNxiii \undefined \def \showISBNxiii  #1{\unskip}     \fi
\ifx \showISSN     \undefined \def \showISSN      #1{\unskip}     \fi
\ifx \showLCCN     \undefined \def \showLCCN      #1{\unskip}     \fi
\ifx \shownote     \undefined \def \shownote      #1{#1}          \fi
\ifx \showarticletitle \undefined \def \showarticletitle #1{#1}   \fi
\ifx \showURL      \undefined \def \showURL       {\relax}        \fi
\providecommand\bibfield[2]{#2}
\providecommand\bibinfo[2]{#2}
\providecommand\natexlab[1]{#1}
\providecommand\showeprint[2][]{arXiv:#2}

\bibitem[Avrahami et~al\mbox{.}(2023)]%
        {avrahami_spatext_2023}
\bibfield{author}{\bibinfo{person}{Omri Avrahami}, \bibinfo{person}{Thomas
  Hayes}, \bibinfo{person}{Oran Gafni}, \bibinfo{person}{Sonal Gupta},
  \bibinfo{person}{Yaniv Taigman}, \bibinfo{person}{Devi Parikh},
  \bibinfo{person}{Dani Lischinski}, \bibinfo{person}{Ohad Fried}, {and}
  \bibinfo{person}{Xi Yin}.} \bibinfo{year}{2023}\natexlab{}.
\newblock \showarticletitle{Spatext: Spatio-textual representation for
  controllable image generation}. In \bibinfo{booktitle}{\emph{Proceedings of
  the IEEE/CVF Conference on Computer Vision and Pattern Recognition}}.
  \bibinfo{pages}{18370--18380}.
\newblock


\bibitem[Avrahami et~al\mbox{.}(2022)]%
        {avrahami2022blended}
\bibfield{author}{\bibinfo{person}{Omri Avrahami}, \bibinfo{person}{Dani
  Lischinski}, {and} \bibinfo{person}{Ohad Fried}.}
  \bibinfo{year}{2022}\natexlab{}.
\newblock \showarticletitle{Blended diffusion for text-driven editing of
  natural images}. In \bibinfo{booktitle}{\emph{Proceedings of the IEEE/CVF
  Conference on Computer Vision and Pattern Recognition}}.
  \bibinfo{pages}{18208--18218}.
\newblock


\bibitem[Balaji et~al\mbox{.}(2022)]%
        {balaji_ediff-i_2023}
\bibfield{author}{\bibinfo{person}{Yogesh Balaji}, \bibinfo{person}{Seungjun
  Nah}, \bibinfo{person}{Xun Huang}, \bibinfo{person}{Arash Vahdat},
  \bibinfo{person}{Jiaming Song}, \bibinfo{person}{Karsten Kreis},
  \bibinfo{person}{Miika Aittala}, \bibinfo{person}{Timo Aila},
  \bibinfo{person}{Samuli Laine}, \bibinfo{person}{Bryan Catanzaro},
  {et~al\mbox{.}}} \bibinfo{year}{2022}\natexlab{}.
\newblock \showarticletitle{ediffi: Text-to-image diffusion models with an
  ensemble of expert denoisers}.
\newblock \bibinfo{journal}{\emph{arXiv preprint arXiv:2211.01324}}
  (\bibinfo{year}{2022}).
\newblock


\bibitem[Bar-Tal et~al\mbox{.}(2023)]%
        {bar-tal_multidiffusion_2023}
\bibfield{author}{\bibinfo{person}{Omer Bar-Tal}, \bibinfo{person}{Lior Yariv},
  \bibinfo{person}{Yaron Lipman}, {and} \bibinfo{person}{Tali Dekel}.}
  \bibinfo{year}{2023}\natexlab{}.
\newblock \showarticletitle{Multidiffusion: Fusing diffusion paths for
  controlled image generation}.
\newblock \bibinfo{journal}{\emph{arXiv preprint arXiv:2302.08113}}
  \bibinfo{volume}{2} (\bibinfo{year}{2023}).
\newblock


\bibitem[Chefer et~al\mbox{.}(2023)]%
        {chefer2023attend}
\bibfield{author}{\bibinfo{person}{Hila Chefer}, \bibinfo{person}{Yuval
  Alaluf}, \bibinfo{person}{Yael Vinker}, \bibinfo{person}{Lior Wolf}, {and}
  \bibinfo{person}{Daniel Cohen-Or}.} \bibinfo{year}{2023}\natexlab{}.
\newblock \showarticletitle{Attend-and-excite: Attention-based semantic
  guidance for text-to-image diffusion models}.
\newblock \bibinfo{journal}{\emph{arXiv preprint arXiv:2301.13826}}
  (\bibinfo{year}{2023}).
\newblock


\bibitem[Gal et~al\mbox{.}(2022)]%
        {gal_image_2022}
\bibfield{author}{\bibinfo{person}{Rinon Gal}, \bibinfo{person}{Yuval Alaluf},
  \bibinfo{person}{Yuval Atzmon}, \bibinfo{person}{Or Patashnik},
  \bibinfo{person}{Amit~H Bermano}, \bibinfo{person}{Gal Chechik}, {and}
  \bibinfo{person}{Daniel Cohen-Or}.} \bibinfo{year}{2022}\natexlab{}.
\newblock \showarticletitle{An image is worth one word: Personalizing
  text-to-image generation using textual inversion}.
\newblock \bibinfo{journal}{\emph{arXiv preprint arXiv:2208.01618}}
  (\bibinfo{year}{2022}).
\newblock


\bibitem[Hertz et~al\mbox{.}(2022)]%
        {hertz_prompt--prompt_2022}
\bibfield{author}{\bibinfo{person}{Amir Hertz}, \bibinfo{person}{Ron Mokady},
  \bibinfo{person}{Jay Tenenbaum}, \bibinfo{person}{Kfir Aberman},
  \bibinfo{person}{Yael Pritch}, {and} \bibinfo{person}{Daniel Cohen-Or}.}
  \bibinfo{year}{2022}\natexlab{}.
\newblock \showarticletitle{Prompt-to-prompt image editing with cross attention
  control}.
\newblock \bibinfo{journal}{\emph{arXiv preprint arXiv:2208.01626}}
  (\bibinfo{year}{2022}).
\newblock


\bibitem[Ho et~al\mbox{.}(2020)]%
        {ho_denoising_2020}
\bibfield{author}{\bibinfo{person}{Jonathan Ho}, \bibinfo{person}{Ajay Jain},
  {and} \bibinfo{person}{Pieter Abbeel}.} \bibinfo{year}{2020}\natexlab{}.
\newblock \showarticletitle{Denoising diffusion probabilistic models}.
\newblock \bibinfo{journal}{\emph{Advances in Neural Information Processing
  Systems}}  \bibinfo{volume}{33} (\bibinfo{year}{2020}),
  \bibinfo{pages}{6840--6851}.
\newblock


\bibitem[Kumari et~al\mbox{.}(2023)]%
        {kumari_multi-concept_2022}
\bibfield{author}{\bibinfo{person}{Nupur Kumari}, \bibinfo{person}{Bingliang
  Zhang}, \bibinfo{person}{Richard Zhang}, \bibinfo{person}{Eli Shechtman},
  {and} \bibinfo{person}{Jun-Yan Zhu}.} \bibinfo{year}{2023}\natexlab{}.
\newblock \showarticletitle{Multi-concept customization of text-to-image
  diffusion}. In \bibinfo{booktitle}{\emph{Proceedings of the IEEE/CVF
  Conference on Computer Vision and Pattern Recognition}}.
  \bibinfo{pages}{1931--1941}.
\newblock


\bibitem[Li et~al\mbox{.}(2023)]%
        {li_gligen_2023}
\bibfield{author}{\bibinfo{person}{Yuheng Li}, \bibinfo{person}{Haotian Liu},
  \bibinfo{person}{Qingyang Wu}, \bibinfo{person}{Fangzhou Mu},
  \bibinfo{person}{Jianwei Yang}, \bibinfo{person}{Jianfeng Gao},
  \bibinfo{person}{Chunyuan Li}, {and} \bibinfo{person}{Yong~Jae Lee}.}
  \bibinfo{year}{2023}\natexlab{}.
\newblock \showarticletitle{GLIGEN: Open-Set Grounded Text-to-Image
  Generation}.
\newblock \bibinfo{journal}{\emph{CVPR}} (\bibinfo{year}{2023}).
\newblock


\bibitem[L{\"u}ddecke and Ecker(2022)]%
        {luddecke2022image}
\bibfield{author}{\bibinfo{person}{Timo L{\"u}ddecke} {and}
  \bibinfo{person}{Alexander Ecker}.} \bibinfo{year}{2022}\natexlab{}.
\newblock \showarticletitle{Image segmentation using text and image prompts}.
  In \bibinfo{booktitle}{\emph{Proceedings of the IEEE/CVF Conference on
  Computer Vision and Pattern Recognition}}. \bibinfo{pages}{7086--7096}.
\newblock


\bibitem[Nichol et~al\mbox{.}(2021)]%
        {nichol_glide_2022}
\bibfield{author}{\bibinfo{person}{Alex Nichol}, \bibinfo{person}{Prafulla
  Dhariwal}, \bibinfo{person}{Aditya Ramesh}, \bibinfo{person}{Pranav Shyam},
  \bibinfo{person}{Pamela Mishkin}, \bibinfo{person}{Bob McGrew},
  \bibinfo{person}{Ilya Sutskever}, {and} \bibinfo{person}{Mark Chen}.}
  \bibinfo{year}{2021}\natexlab{}.
\newblock \showarticletitle{Glide: Towards photorealistic image generation and
  editing with text-guided diffusion models}.
\newblock \bibinfo{journal}{\emph{arXiv preprint arXiv:2112.10741}}
  (\bibinfo{year}{2021}).
\newblock


\bibitem[Ramesh et~al\mbox{.}(2022)]%
        {ramesh2022hierarchical}
\bibfield{author}{\bibinfo{person}{Aditya Ramesh}, \bibinfo{person}{Prafulla
  Dhariwal}, \bibinfo{person}{Alex Nichol}, \bibinfo{person}{Casey Chu}, {and}
  \bibinfo{person}{Mark Chen}.} \bibinfo{year}{2022}\natexlab{}.
\newblock \showarticletitle{Hierarchical text-conditional image generation with
  clip latents}.
\newblock \bibinfo{journal}{\emph{arXiv preprint arXiv:2204.06125}}
  (\bibinfo{year}{2022}).
\newblock


\bibitem[Rombach et~al\mbox{.}(2022)]%
        {rombach_high-resolution_2022}
\bibfield{author}{\bibinfo{person}{Robin Rombach}, \bibinfo{person}{Andreas
  Blattmann}, \bibinfo{person}{Dominik Lorenz}, \bibinfo{person}{Patrick
  Esser}, {and} \bibinfo{person}{Bj{\"o}rn Ommer}.}
  \bibinfo{year}{2022}\natexlab{}.
\newblock \showarticletitle{High-resolution image synthesis with latent
  diffusion models}. In \bibinfo{booktitle}{\emph{Proceedings of the IEEE/CVF
  Conference on Computer Vision and Pattern Recognition}}.
  \bibinfo{pages}{10684--10695}.
\newblock


\bibitem[Ruiz et~al\mbox{.}(2023)]%
        {ruiz_dreambooth_2023}
\bibfield{author}{\bibinfo{person}{Nataniel Ruiz}, \bibinfo{person}{Yuanzhen
  Li}, \bibinfo{person}{Varun Jampani}, \bibinfo{person}{Yael Pritch},
  \bibinfo{person}{Michael Rubinstein}, {and} \bibinfo{person}{Kfir Aberman}.}
  \bibinfo{year}{2023}\natexlab{}.
\newblock \showarticletitle{Dreambooth: Fine tuning text-to-image diffusion
  models for subject-driven generation}. In
  \bibinfo{booktitle}{\emph{Proceedings of the IEEE/CVF Conference on Computer
  Vision and Pattern Recognition}}. \bibinfo{pages}{22500--22510}.
\newblock


\bibitem[Saharia et~al\mbox{.}(2022)]%
        {saharia_photorealistic_nodate}
\bibfield{author}{\bibinfo{person}{Chitwan Saharia}, \bibinfo{person}{William
  Chan}, \bibinfo{person}{Saurabh Saxena}, \bibinfo{person}{Lala Li},
  \bibinfo{person}{Jay Whang}, \bibinfo{person}{Emily~L Denton},
  \bibinfo{person}{Kamyar Ghasemipour}, \bibinfo{person}{Raphael
  Gontijo~Lopes}, \bibinfo{person}{Burcu Karagol~Ayan}, \bibinfo{person}{Tim
  Salimans}, {et~al\mbox{.}}} \bibinfo{year}{2022}\natexlab{}.
\newblock \showarticletitle{Photorealistic text-to-image diffusion models with
  deep language understanding}.
\newblock \bibinfo{journal}{\emph{Advances in Neural Information Processing
  Systems}}  \bibinfo{volume}{35} (\bibinfo{year}{2022}),
  \bibinfo{pages}{36479--36494}.
\newblock


\bibitem[Song et~al\mbox{.}(2020)]%
        {song_denoising_2022}
\bibfield{author}{\bibinfo{person}{Jiaming Song}, \bibinfo{person}{Chenlin
  Meng}, {and} \bibinfo{person}{Stefano Ermon}.}
  \bibinfo{year}{2020}\natexlab{}.
\newblock \showarticletitle{Denoising diffusion implicit models}.
\newblock \bibinfo{journal}{\emph{arXiv preprint arXiv:2010.02502}}
  (\bibinfo{year}{2020}).
\newblock


\bibitem[Zhang and Agrawala(2023)]%
        {zhang_adding_2023}
\bibfield{author}{\bibinfo{person}{Lvmin Zhang} {and} \bibinfo{person}{Maneesh
  Agrawala}.} \bibinfo{year}{2023}\natexlab{}.
\newblock \showarticletitle{Adding conditional control to text-to-image
  diffusion models}.
\newblock \bibinfo{journal}{\emph{arXiv preprint arXiv:2302.05543}}
  (\bibinfo{year}{2023}).
\newblock


\end{thebibliography}

\end{document}